\documentclass[letterpaper, 10 pt, conference]{ieeeconf}  

\IEEEoverridecommandlockouts                              

\usepackage{color}
\usepackage{amsmath, amssymb, amscd, amsfonts} 
\usepackage{graphics} 
\usepackage{graphicx}
\usepackage{tabularx}
\usepackage{multirow}
\usepackage{colortbl}
\usepackage{hhline}
\usepackage{booktabs}
\usepackage{lipsum}
\usepackage{cite}
\usepackage{float}
\usepackage{subfig}
\usepackage{caption}
\usepackage{algorithm}
\usepackage{algorithmic}
\usepackage{bbding}
\usepackage{pifont}
\title{\LARGE \bf
Learning Agile Gate Traversal via Analytical Optimal Policy Gradient
}

\author{
    Tianchen Sun, Bingheng Wang, Nuthasith Gerdpratoom,  Longbin Tang, Yichao Gao, Lin Zhao
    \thanks{
        These authors are with the Department of Electrical and Computer Engineering, National           University of Singapore, Singapore. Email: 
        {\tt\small 
            tianchen.sun@u.nus.edu, wangbingheng@u.nus.edu, nuthasith@u.nus.edu, longbin@u.nus.edu, 
            yichao\_gao@u.nus.edu, zhaolin@nus.edu.sg
        }
    }
}

\begin{document}

\maketitle
\thispagestyle{empty}
\pagestyle{empty}

\begin{abstract}

Traversing narrow gates presents a significant challenge and has become a standard benchmark for evaluating agile and precise quadrotor flight. Traditional modularized autonomous flight stacks require extensive design and parameter tuning, while end-to-end reinforcement learning (RL) methods often suffer from low sample efficiency, limited interpretability, and degraded disturbance rejection under unseen perturbations. In this work, we present a novel hybrid framework that adaptively fine-tunes model predictive control (MPC) parameters online using outputs from a neural network (NN) trained offline. The NN jointly predicts a reference pose and cost function weights, conditioned on the coordinates of the gate corners and the current drone state. To achieve efficient training, we derive analytical policy gradients not only for the MPC module but also for an optimization-based gate traversal detection module. Hardware experiments demonstrate agile and accurate gate traversal with peak accelerations of $30\ \mathrm{m/s^2}$, as well as recovery within $0.85\ \mathrm{s}$ following body-rate disturbances exceeding $1146\ \mathrm{deg/s}$.

\end{abstract}

\section{INTRODUCTION}

Motion planning and control within constrained spaces for a quadrotor is inherently challenging 
due to its underactuated nature, where the translational and rotational dynamics are coupled. Narrow gate traversal is a representative and demanding task that requires highly agile flight, precise pose control, and strict adherence to spatiotemporal constraints. It is widely employed as a standard benchmark for evaluating agile motion planning and control methods~\cite{neunert2016fast,falanga2017aggressive,WANG2022GCOPTER,wu2025whole}.

Traditional flight control for narrow gate traversal typically follows a hierarchical architecture comprising path planning \cite{liu2018search}, open-loop trajectory generation \cite{WANG2022GCOPTER}, and closed-loop tracking control \cite{lee2010geometric}. While this modular approach facilitates practical development by decomposing the overall task, it generally demands extensive parameter tuning. Moreover, because the upper modules operate at progressively lower frequencies and each module is tuned with static parameters (e.g. fixed weights), the overall framework exhibits limited ability to adapt rapidly to model uncertainties and environmental changes.
\begin{figure}[!t]
    \centering
    \vspace{4mm}
    \includegraphics[width=1.0\linewidth]{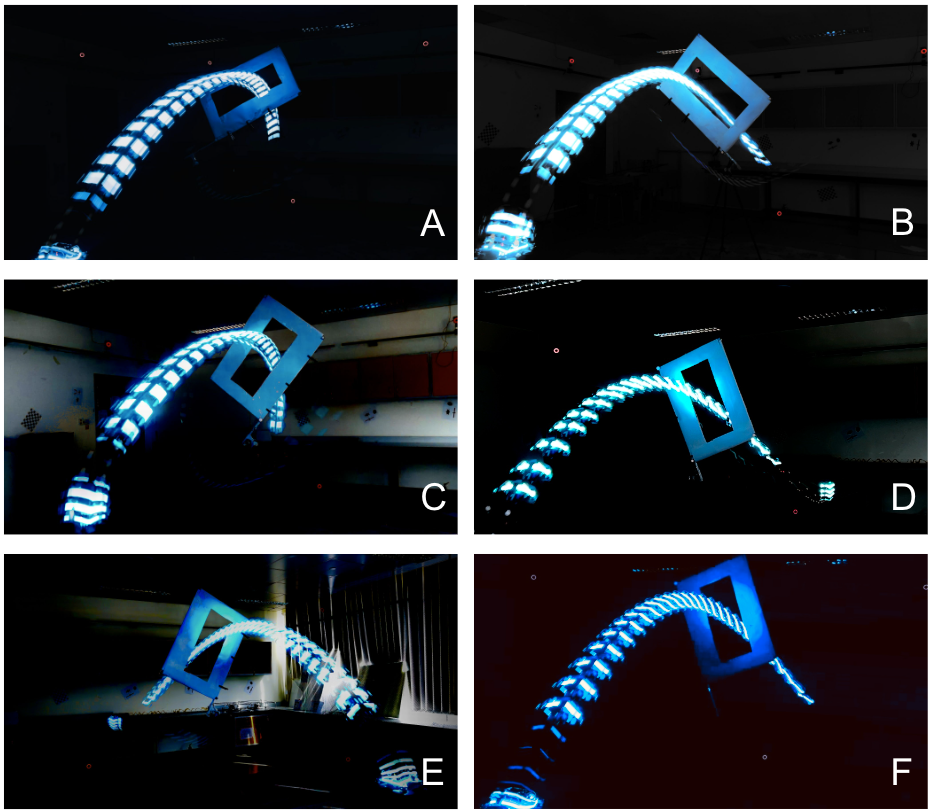}
    \caption{Multiple real flight demonstrations with gate angle ranging from $30^{\circ}$ to $70^{\circ}$. The trajectory of the quadrotor is illustrated through composited images generated from sequential snapshots. The gate orientations from subfigures A to F are $30^{\circ}, \ -45^{\circ}, \ 45^{\circ}, \ -65^{\circ}, \ 60^{\circ}$, and $-70^{\circ}$, respectively.}
    \label{fig: real flight photos}
    \vspace{-0.6cm}
\end{figure}

Recent research has increasingly focused on learning end-to-end neural network policies for quadrotor agile flight control~\cite{wu2025whole,xie2023learning,song2023reaching}. These policies directly map observations to control actions. However, it is generally difficult to enforce additional hard constraints within such frameworks, and because the training typically does not utilize prior knowledge such as system dynamics, sample efficiency is poor. Moreover, in the absence of online optimization, the robustness of neural network policies largely depends on domain randomization during offline training \cite{hu2025narrow,ferede2025one}, which may result in degraded disturbance rejection under unseen and extreme perturbations.

To leverage the strengths of both model-based and model-free approaches, methods that combine MPC and NN have been proposed in~\cite{song2022policy,romero2025actor,wang2023learning}.
Song et al.~\cite{song2022policy} train an NN to predict the time a quadrotor requires to traverse a gate. The predicted time is then used to adjust the gate tracking weight in the MPC cost function. They employ a two-stage training that involves a Gaussian linear policy search followed by imitation learning. However, the learning process entails a large number of MPC evaluations, making it computationally expensive and sample inefficient. 
Romero et al.\cite{romero2025actor} apply an actor–critic framework, trained with the PPO algorithm to tune the weights of a generic quadratic cost function in MPC using the actor network, with training facilitated by differentiable MPC~\cite{amos2018differentiable}. Nevertheless, the overall policy gradient is still estimated from sampled trajectories, resulting in high variance and noisy updates.
Wang et al.\cite{wang2023learning} train an NN to predict both the traversal time and a six-dimensional reference pose to guide a quadrotor through gates. The NN is trained by maximizing a discrete reward for successful traversal. They treat MPC as a black box and use finite differences to approximate the policy gradient, which significantly improves training efficiency compared to \cite{song2022policy}. However, it remains sample inefficient due to the reliance on numerically approximated gradients.

The aforementioned methods all rely on various forms of numerical gradient approximations, such as Gaussian policy search~\cite{song2022policy}, model-free reinforcement learning sampling~\cite{romero2025actor}, or finite differences~\cite{wang2023learning}. In contrast, we propose a fully differentiable NN–MPC hybrid framework that leverages analytical gradients for efficient learning. Specifically, we formulate the narrow gate traversal task as a reference-pose tracking problem using MPC, where the reference pose is designed to guide the quadrotor through the gate. The NN predicts both the reference poses and the corresponding MPC cost-term weights in real time, enabling fast online adaptation. Fig.~\ref{fig: title_fig} shows the NN-predicted poses and the corresponding MPC-predicted trajectories at $t=0.72\text{s},\ 0.84\text{s},\ 0.96\text{s}$ from a real flight experiment. The corresponding traversal trajectory is captured in Fig.~\ref{fig: real flight photos} D.

To enable efficient and stable training, we introduce a formulation of the attitude error in the MPC cost that employs an unconstrained 3$\times$3 matrix as the attitude reference. In contrast, the three-dimensional Rodrigues parameter rotation representation adopted by Wang et al.\cite{wang2023learning} inherently induces discontinuous gradients\cite{geist2024learning}, which can severely degrade learning performance. Furthermore, we formulate the gate collision detection as a differentiable conic optimization problem, which is also generalizable to different scenarios. By analytically differentiating through both the MPC and the conic optimization, the training efficiency is improved. We term the resulting gradient for updating the neural networks, derived from these differentiable optimizations, the \textit{analytical optimal policy gradient}. Fig.~\ref{fig: overview_pipeline} illustrates the block diagram of our proposed NN–MPC framework and its learning pipeline.

Our contributions are summarized as follows:
\begin{enumerate}
\item We develop a fully differentiable NN-MPC with learnable time-varying cost weights and a single reference pose for agile and accurate quadrotor gate traversal in confined spaces. It enables adaptive objective emphasis online and can be efficiently trained offline with faster gradient computation.
 
\item Our framework realizes zero-shot sim-to-real transfer and maintains effective disturbance rejection, as it preserves the online optimization feature of MPC.

\item We demonstrate our approach on challenging narrow-gate traversal tasks through extensive simulation and real-world experiments. Hardware experiments demonstrate that our method not only achieves accurate and agile gate traversal with peak acceleration of $30\ \mathrm{m/s^2}$, but also enables rapid recovery within $0.85\ \mathrm{s}$ from extreme body-rate disturbances exceeding $1146\ \mathrm{deg/s}$.

\end{enumerate}
The remainder of this paper is organized as follows. Section~\ref{sec: MPC formulation} presents the problem formulation. Section~\ref{sec: APG} derives the analytical policy gradient. Section~\ref{sec: experiments} reports our simulation and experimental results with benchmark comparisons.

\section{Problem Formulation} \label{sec: MPC formulation}
\subsection{Quadrotor Dynamics} \label{sec: MPC intro}
The continuous-time quadrotor dynamics are are given as follows, where we employ collective thrust and body-rate (CTBR) control:
\begin{subequations}
\begin{align} 
\label{func: quadrotor}
^{\mathcal{W}}\dot{\mathbf{p}} & =^{\mathcal{W}}\mathbf{v}, \\ 
^{\mathcal{W}} \dot{\mathbf{v}}& =m^{-1} \mathbf{R}(\mathbf{q}) f_{\mathrm{r}} \mathbf{e}_{z}-g \mathbf{e}_{z}, \\ 
\dot{\mathbf{q}} & =\frac{1}{2}\boldsymbol{\Omega}\left(\boldsymbol{^{\mathcal{B}}\omega}\right) \mathbf{q},
\end{align}
\end{subequations}
where $^{\mathcal{W}} \mathbf{p}=[x,y,z]^T, ^{\mathcal{W}}\mathbf{v}=[v_x,v_y,v_z]^T$ are the quadrotor position and velocity vector in the world frame $\mathcal{W}$, respectively. $\mathbf{q}=[q_0,q_x,q_y,q_z]^T$ is the quaternion from the body frame $\mathcal{B}$ to the world frame $\mathcal{W}$, $m$ is the mass of the quadrotor, $\mathbf{R}(\mathbf{q})$ is the rotation matrix corresponding to $\mathbf{q}$, $\boldsymbol{^{\mathcal{B}}{\boldsymbol{\omega}}}=[\omega_x,\omega_y,\omega_z]^T$ is the quadrotor body rate expressed in the body frame, $\boldsymbol{\Omega}(^{\mathcal{B}}\boldsymbol{\omega})$ is the corresponding skew-symmetric matrix,  $f_r$ is the collective thrust, $e_z=[0,0,1]^T$ is a basis vector, and $g=9.81 \  \mathrm{m/s^2}$ is a gravitational acceleration. 

\subsection{Reference Tracking MPC} \label{sec:cost design} 
We formulate a reference tracking MPC for the gate traversal task. Over a finite horizon $N$, the MPC solves the following optimal control problem (OCP):
\begin{align}
\label{func: general_MPC}
\begin{array}{rl}
\pi(\mathbf{x}) = \underset{\mathbf{\xi}}{\operatorname{argmin}} & 
J(\mathbf{\xi}) = \sum_{k=0}^{N-1} c(\mathbf{x}_k, \mathbf{u}_k) + c_N(\mathbf{x}_N) \\[0.5em]
\text{s.t.}
& \mathbf{x}_{k+1} = f(\mathbf{x}_k, \mathbf{u}_k), \\ 
& g(\mathbf{x}_k, \mathbf{u}_k) \leq 0,\quad g_N(\mathbf{x}_N) \leq 0, \\
& \mathbf{x}_0 = \mathbf{x}_{\text{init}},\notag
\end{array}
\end{align}
where $f$ is the discrete-time dynamics, and $g, g_N$ represent state and input constraints. This OCP is solved at each time step to obtain a predicted optimal state-control sequence $\mathbf{\xi}:=\{\mathbf{x}_{0:N},\mathbf{u}_{0:N-1}\}$, and only the first control $\mathbf{u}_0$ is applied to the system. The dynamics are discretized using a fourth-order Runge--Kutta method. The quadrotor state is denoted by $\mathbf{x} = [\mathbf{p}, \mathbf{v}, \mathbf{q}]^\top$, 
and the control is $\mathbf{u} = [f_r, {^{\mathcal{B}}\boldsymbol{\omega}}]^\top$.

\begin{figure*}[!t]
    \centering
    \vspace{4mm}
    \includegraphics[width=0.88\textwidth]{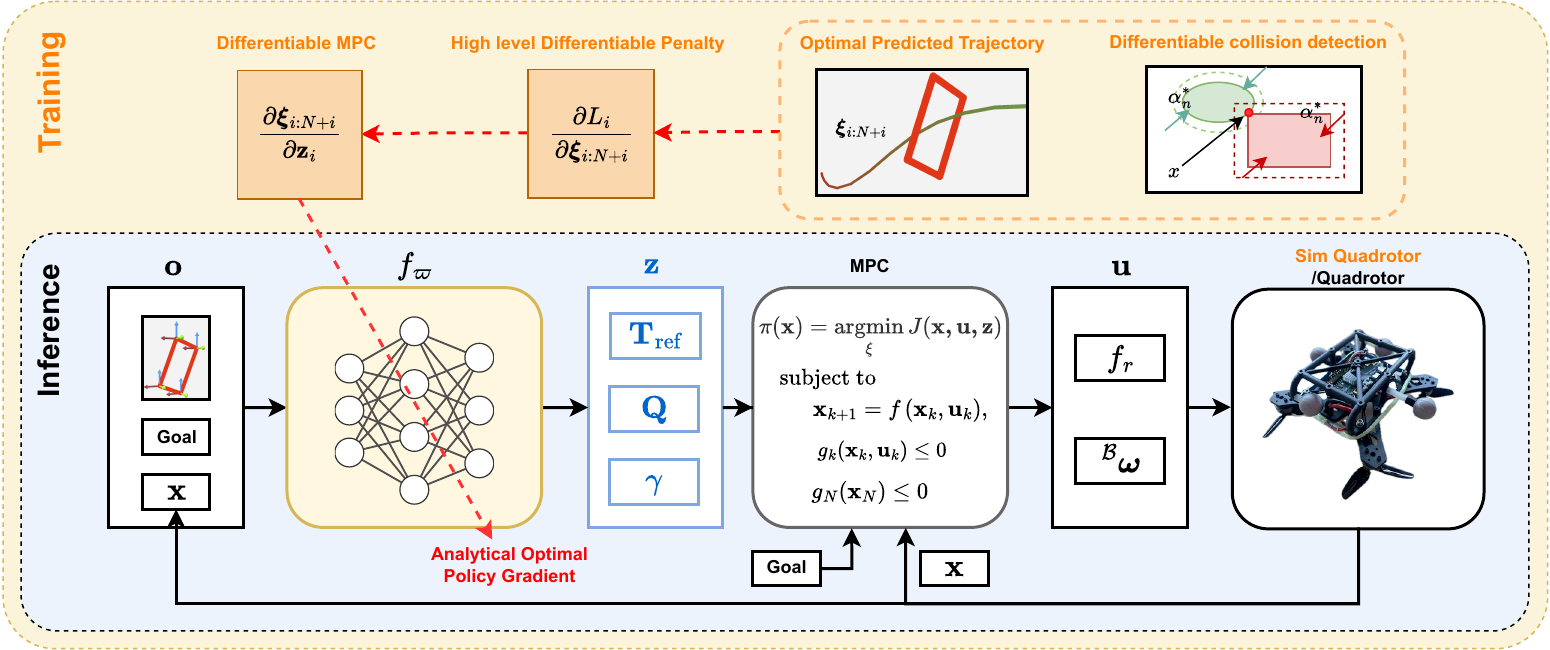}
    \caption{In this framework, the inference pass features two nested closed-loop feedback structures. In the outer closed loop, a NN predicts both the reference pose $\mathbf{T}_{\text{ref}}$ and the cost-terms weights $\mathbf{Q}$, $\gamma$, based on the observed gate corner positions, the goal position, and the current state of the quadrotor $\mathbf{x}$. In the inner closed loop, the MPC module optimizes the predicted state and control trajectory, $\mathbf{\xi}_i$, and only the first control input, $\mathbf{u} = [f_r, {^{\mathcal{B}}{{\boldsymbol{\omega}}}}]$, is applied to the quadrotor.
    During training, a high-level loss is imposed by constructing a differentiable gate collision detection  
    problem to evaluate the MPC predicted trajectory. By composing the gradients from both the differentiable MPC and the differentiable gate collision detection module, an analytical optimal policy gradient is obtained to update the NN.}
    \label{fig: overview_pipeline}
    \vspace{-0.4cm}
\end{figure*}

The MPC stage cost $c$ contains the reference pose tracking cost $c_{\mathbf{p}}$, $c_{\mathbf{R}}$, the goal-reaching cost $c_{\text{goal}}$, and the control regulation cost $c_{\mathbf{u}}$:
\begin{align}
\begin{array}{ll}
c(\mathbf{x}_{k} , \mathbf{u}_{k})= & c_{\mathbf{p}}(\mathbf{p}_k,\mathbf{p}_{\text{ref}})+ 
c_{\mathbf{R}}(\mathbf{q}_k,\mathbf{R}_{\text{ref}}) \\ &+
c_{\text{goal}}(\mathbf{x}_k,\mathbf{x}_{\text{goal}})+
c_{\mathbf{u}} (\mathbf{u}_k),
\end{array}
\end{align}
where $[\mathbf{p}_{\text{ref}},\mathbf{R}_{\text{ref}}]:=\mathbf{T}_{\text{ref}}$ is the reference pose.
The position tracking cost is defined using the weighted 2-norm with the time-varying weight $\tilde{\mathbf{Q}}_{\mathbf{p}_{\text{ref}}}$:
\begin{align}
\label{func: c_{posi}}
c_{\mathbf{p}}(\mathbf{p},\mathbf{p}_{\text{ref}})=(\mathbf{p}-\mathbf{p}_{\text{ref}})^T{\tilde{\mathbf{Q}}_{\mathbf{p}_{\text{ref}}}}(\mathbf{p}-\mathbf{p}_{\text{ref}}).
\end{align}
The attitude tracking cost is defined using the Frobenius norm between the reference attitude $\mathbf{R}_{\text{ref}}$ and the MPC-predicted quadrotor attitude $\mathbf{R}(\mathbf{q})$, which is:
\begin{align}
\label{func: c_{ori}}
c_{\mathbf{R}}(\mathbf{q},\mathbf{R}_{\text{ref}})= \tilde{\mathbf{Q}}_{\mathbf{R}_{\text{ref}}}  \|\mathbf{R}(\mathbf{q})-\mathbf{R}_{\text{ref}}\|_{\mathrm{F}}^2,
\end{align}
where $\tilde{\mathbf{Q}}_{\mathbf{R}_{\text{ref}}}$ is a time-varying scalar weight.
The goal-reaching cost and the control regulation cost are:
\begin{align}
\label{func: c_{goal}}
c_{\text{goal}}(\mathbf{x} ,\mathbf{x}_{\text{goal}})= (\mathbf{x} -\mathbf{x}_{\text{goal}})^T{\mathbf{Q}_{\mathbf{x}_{\text{goal}}}}(\mathbf{x} -\mathbf{x}_{\text{goal}}), \\
c_{\mathbf{u}} (\mathbf{u})=(\mathbf{u}-\mathbf{u}_{\text{hover}})^T{\mathbf{Q}_{\mathbf{u}}}(\mathbf{u}-\mathbf{u}_{\text{hover}}),
\end{align}
respectively, where $\mathbf{Q}_{\mathbf{x}_{\text{goal}}}:=\text{diag}(\tilde{\mathbf{Q}}_{\mathbf{p}_{\text{goal}}},\mathbf{Q}_{\mathbf{v}_{\text{goal}}}, \mathbf{Q}_{\mathbf{q}_{\text{goal}}} ) \in \mathbb{R}^{10\times 10}$ is a diagonal matrix, and $\mathbf{u}_\text{hover}$ denotes the input required to keep the quadrotor in hover. The terminal cost $c_N(\mathbf{x}_N)$ only contains the goal reaching cost $c_{\text{goal}}(\mathbf{x}_N,\mathbf{x}_{\text{goal}})$. 

We adopt diagonal matrices $\tilde{\mathbf{Q}}_{\mathbf{p}_{\text{ref}}}$, $\tilde{\mathbf{Q}}_{\mathbf{p}_{\text{goal}}} \in \mathbb{R}^{3\times3}$ for the weights of the reference position-tracking and goal position-reaching costs, respectively. 
As the quadrotor must traverse the gate before reaching the goal, we further consider the following time-varying weights to adaptively prioritize different cost terms over the prediction horizon:
\begin{subequations}
\begin{align}
\label{eq:c_{posi_vary}}
\tilde{\mathbf{Q}}_{\mathbf{p}_{\text{ref}}}&=\frac{1}{2}\mathbf{Q}_{\mathbf{p}_{\text{ref}}}(1+\text{tanh}(1000(t_{\text{ref}}-k d_{t}))),
\\
\label{eq:c_{goal_vary}}
\tilde{\mathbf{Q}}_{\mathbf{p}_{\text{goal}}}&=\frac{1}{2}\mathbf{Q}_{\mathbf{p}_{\text{goal}}}(1+\text{tanh}(1000(k d_{t}-t_\text{ref}))),
\\
\label{eq:c_{ori_vary}}
\tilde{\mathbf{Q}}_{\mathbf{R}_{\text{ref}}} &=\mathbf{Q}_{\mathbf{R}_{\text{ref}}} \exp \left(-\gamma \left(k  d_{t}-t_{\text{ref}}\right)^{2}\right),
\end{align}
\end{subequations} 
where $d_t$ is the MPC sampling interval and $k\in\{0,1,2,\dots,N-1\}$ the prediction time index. $t_{\text{ref}}$ denotes the reference gate traversal time, which is fixed and estimated from the gate distance and the desired velocity. From \eqref{eq:c_{posi_vary}} and \eqref{eq:c_{goal_vary}}, it can be observed that the relative priority between reference tracking and goal reaching transitions around $t_{\text{ref}}$. The exponential term in the function (\ref{eq:c_{ori_vary}}) forms a bell-shaped curve centered around $t_{\text{ref}}$, with $\gamma$ controlling the temporal sharpness of the weight. 

We will train an NN to generate a set of high-level decision variables consisting of the reference pose $\mathbf{T}_{\text{ref}}$, 
the diagonal elements of $\mathbf{Q}_{\mathbf{p}_{\text{ref}}}$ and $\mathbf{Q}_{\mathbf{p}_{\text{goal}}}$, 
as well as the scalar weights $\mathbf{Q}_{\mathbf{R}_{\text{ref}}}$ and $\gamma$.
These variables are summarized as $\mathbf{z} \in \mathbb{R}^{20}$:
\begin{align}
\label{func: z define}
    \mathbf{z} =
    \big[
    \mathbf{p}_{\text{ref}},\,
    \mathbf{R}_{\text{ref}},\,
    \text{diag}\!\left(\mathbf{Q}_{\mathbf{p}_{\text{ref}}}\right),\,
    \text{diag}\!\left(\mathbf{Q}_{\mathbf{p}_{\text{goal}}}\right),\,
    \mathbf{Q}_{\mathbf{R}_{\text{ref}}},\,
    \gamma
    \big].
\end{align}

\subsection{Attitude Error Representation} \label{sec: rotation rep}
Learning rotation is non-trivial, as different representations of rotation can significantly affect the learning performance~\cite{geist2024learning}. The major reason lies in the ill-posedness of rotation learning due to discontinuities and singularities in common rotation representations. Empirical studies show that an unconstrained intermediate $3\times3$  matrix $\mathbf{M_{\text{ref}}} \in \mathbb{R}^{3\times 3}$ representation is, in general, a more expressive and numerically stable parameterization of rotations~\cite{SVD}. The corresponding rotation can be recovered from the projection of $\mathbf{M_{\text{ref}}}$ to the closest rotation matrix via singular value decomposition (SVD). Mathematically, the rotation matrix is computed by 
\begin{align}
\label{func: svd}
 \mathbf{R}_{\text{ref}}=\operatorname{SVD}^+(\mathbf{M_{\text{ref}}}):=U \text{diag}(1,1, \det(UV^T))V^T,
\end{align}
where the superscript $+$ denotes that a rotation matrix with determinant $1$ is to be obtained by correcting the sign using $\det(UV^T)$.

By using this representation in the MPC cost function~\eqref{func: c_{ori}}, the attitude tracking error is computed by:
\begin{align}
\label{func: c_{ori} with svd}
\|\mathbf{R}_k(\mathbf{q}_k)-\operatorname{SVD^+}(\mathbf{M}_{\text{ref}})\|^2_{\mathrm{F}},
\end{align}
This rotation error formulated in the Frobenius norm is equivalent to an axis–angle representation \cite{hartley2013rotation}, and remains effective for both small and large angular errors.

\subsection{High-level Loss}
\label{sec:penalty design}
The high-level loss for the task consists of three components: the gate traversal loss $L_{\text{gate}}$, the goal position-reaching loss $L_{\text{goal}}$, and the control difference loss $L_{\text{control}}$: 
\begin{align}
\label{prob: high level loss composite}
L=L_{\text{gate}}+L_{\text{goal}}+L_{\text{control}}.
\end{align}
Specifically, for $L_{\text{gate}}$, we formulate the gate collision detection as a differentiable conic optimization problem \cite{tracy2023differentiable}. This problem minimizes a scaling factor $\alpha$ at which two convex objects come into contact. We represent the gate boundary as four polytopes, denoted as $\mathcal{C}_{\text{gate}\_n}(\mathbf{T}_{\text{gate}\_n},\alpha_n), n \in \{0,1,2,3\}$, parameterized by the pose $\mathbf{T}_{\text{gate}\_n}=[\mathbf{p}_{\text{gate}\_n},\mathbf{q}_{\text{gate}\_n}]^T$. The quadrotor is commonly represented as an ellipsoid~\cite{WANG2022GCOPTER}, which admits a formulation as a second-order cone constraint. Here we denote it by $\mathcal{C}_{\text{quad}}(\mathbf{T}_{\text{quad}},\alpha_n)$, where the quadrotor pose $\mathbf{T}_{\text{quad}} = [\mathbf{p}_k, \mathbf{q}_k]^T$ is taken from the $\mathbf{\xi}$ at time instances in which the distance to the gate position is within a predefined threshold. At each of these times $k$, using the convex body representations above, we can formulate the collision detection between the drone and the gate elements as the following conic optimization problem:
\begin{align}
\label{prob:diff collision}
\begin{array}{ll}
\underset{x, \alpha_n}{\operatorname{min}} & \alpha_n=f(\alpha_n) \\ 
\text { s.t.  } & x \in \mathcal{C}_{\text{quad}}(\mathbf{T}_{\text{quad}},\alpha_n), \\
& x \in \mathcal{C}_{\text{gate}\_n}(\mathbf{T}_{\text{gate}\_n},\alpha_n),
\\ & \alpha_n \geq 0,
\end{array}
\end{align}
for each $n \in \{0,1,2,3\}$, where $x$ is a spatial point, and its optimal value $x^*$ is the contact point. The optimal solution of problem (\ref{prob:diff collision}) is the minimum scaling $\alpha^*_n \geq 0$ such that the point $x^*$ lies within both convex sets $\mathcal{C}_{\text{quad}}(\mathbf{T}_{\text{quad}},\alpha^*_n)$ and $\mathcal{C}_{\text{gate}\_n}(\mathbf{T}_{\text{gate}},\alpha^*_n)$, which are uniformly scaled by $\alpha^*_n$. The overall optimization process is illustrated in Fig.~\ref{fig: overview_pipeline}, the differentiable collision detection section. By solving problem (\ref{prob:diff collision}), we map the binary collision detection event to a continuous quantity, namely the minimum-scaling factor $\alpha^*_n$:
\begin{align}
\text{collision}=\left\{\begin{array}{l}
\text{True},\  \alpha^*_n \leq 1, \text{ shrink 
 to contact},  \\ 
\text{False}, \  \alpha^*_n > 1,  \text{ enlarge to contact}.
\end{array} \right.
\end{align} 
 
The high-level loss for training the NN should encourage a collision-free gate traversal, which can be translated into requiring $\alpha_n^*>1$ but not excessively too large, as it would cause the quadrotor to deviate far from the gate. To this end, we first inflate the drone ellipsoid so that it exactly matches the gate size (e.g., by scaling its height to equal the gate width), and then define the gate traversal loss as
\begin{align}
\label{func: L_GATE}
    L_{\text{gate}}=\beta_{\text{gate}}\sum_{n=0}^{p-1} (\alpha^*_n-1)^2,
\end{align}
where $p$ is the number of polytopes that consist of the gate, $\beta_{\text{gate}}>0$ is a scalar weight. Minimizing $L_{\text{gate}}$ encourages the quadrotor to pass through the gate with the largest possible safety margin. This method is also generalizable to different scenarios.

We defer the details of differentiation through this optimization to Section~\ref{sec: diff col}.

We formulate the goal position-reaching loss $L_{\text{goal}}$ to penalize the squared Euclidean distances between the last $h$ predicted nodes and the goal position:
\begin{align}
\label{func: L_GOAL}
    L_{\text{goal}}=\beta_{\text{goal}}\sum_{j=N-h+1}^{N}\|\mathbf{p}_j-\mathbf{p}_{\text{goal}}\|^2,
\end{align}
where $h$ and $\beta_{\text{goal}}$ are hyperparameters. 

We encourage the control smoothness by the following term:
\begin{align}
\label{func: L_CONTROL_DIFF}
    L_{\text{control}}=\beta_{\text{control}}\|\mathbf{u}_{i}-\mathbf{u}_{i-1}\|^2,
\end{align}
where $i$ is the actual state timestep. 

Traditionally, penalizing the control difference in MPC often requires augmenting the quadrotor state vector (e.g., to penalize the body rate difference, the state in the MPC formulation is augmented with the body rate and the body rate difference is assigned as the MPC control) \cite{romero2022model}, which introduces additional computational overhead online.

\subsection{Bilevel Optimization Formulation}

We further propose a bilevel optimization framework to minimize the aforementioned loss functions. Specifically, at each actual state timestep $i \in \{0,..,H\}$, with the episode length $H$, the neural network $f_{\mathbf\varpi}$ predicts the optimal high-level decision variable $\mathbf{z}_i$, MPC computes the optimal predicted state-action trajectory $\mathbf{\xi}_i$, and the high-level loss function $L_i$ is calculated as detailed in the last subsection. The objective is to minimize the \textit{cumulative} loss  $\mathbf{L}=\sum^H_{i=0} L_{i}$ over the entire episode horizon, which can be formulated by:
 \begin{align}
 \label{func: bilevel opt}
    \underset{\varpi}{\operatorname{min}}\ &   \mathbf{L}(\boldsymbol{\xi}) \notag 
    \\ 
    \text { s.t.  }   \boldsymbol{\xi}& = \text{MPC}(\mathbf{Z}(\varpi)),
 \end{align}
where $\mathbf{Z}=\{\mathbf{z}_{0:H}\}$, $\boldsymbol{\xi}= \{\mathbf{\xi}_{0:H}\}$ denote the sequence of high-level decision variables and the MPC predicted optimal trajectories over the entire episode, respectively. 
 
\section{Analytical Optimal Policy Gradient} \label{sec: APG}
Our approach optimizes the NN using the analytical optimal policy gradient $\frac{d \mathbf{L}}{d \mathbf\varpi}$. Its component at time $i$ can be computed via the following chain rule:  
\begin{align}
\label{func: current APG}
\frac{d  L_{i}}{d \boldsymbol{\varpi}}
&= 
\underbrace{\frac{\partial L_i}{\partial \mathbf{\xi}_{i}}}_{\substack{\text{diff. through}\\\text{high-level loss}}}
\cdot
\underbrace{\frac{\partial \mathbf{\xi}_{i}}{\partial \mathbf{z}_i}}_{\substack{\text{diff. through MPC}}}
\cdot 
\underbrace{\frac{\partial \mathbf{z}_i}{\partial  \boldsymbol{\varpi}}.}_{\substack{\text{diff. through NN}}}
\end{align}
Here, $\frac{\partial \mathbf{z}_i}{\partial  \mathbf\varpi}$  is the gradient of the neural networks, which is readily available via automatic differentiation (AD). However, the gradient of MPC and collision detection involves differentiation through optimization problems; in particular, the MPC case constitutes a dynamic optimization problem, which is more challenging. Unlike the NN case, where the explicit function of NN is available, these gradients are typically computed via implicit differentiation through the KKT conditions. We detail the computation of these gradients in the following subsection, and the full training algorithm is shown in Algorithm \ref{alg:close loop train}.

\subsection{Differentiate through Collision Detection}\label{sec: diff col}

We first differentiate through the collision detection problem formulated in (\ref{prob:diff collision}). In the forward pass, the collision detection problem is solved using ECOS solver \cite{bib:Domahidi2013ecos}, which yields the optimal solution $x^*,\alpha^*_n$ and $\boldsymbol{\lambda^*}$, where $ \boldsymbol{\lambda}^*$ is the optimal Lagrange multiplier. In the backward pass, we obtain the desired gradient of the minimum scaling with respect to the quadrotor pose $\frac{\partial\alpha^*_n}{\partial \mathbf{T}_{\text{quad}}}$.
We abbreviate decision variables as $\mathbf{y}=[x, \alpha_n]^T $, the problem parameter as $\boldsymbol{\theta}= [\mathbf{T}_{\text{quad}},\mathbf{T}_{\text{gate}\_n}]^T$, and the inequality constraint as $g(\mathbf{y},\boldsymbol{\theta}) \leq 0$. Then
problem (\ref{prob:diff collision}) is expressed as:
\begin{align}
    \begin{array}{ll}
        V(\boldsymbol{\theta})&=\underset{\mathbf{y}}{\operatorname{min}}  f(\mathbf{y}(\boldsymbol{\theta})) \\ 
        \text { s.t.  } & g(\mathbf{y},\boldsymbol{\theta}) \leq 0,
    \end{array}
\end{align}
where $V(\boldsymbol{\theta})=f(\mathbf{y}^*(\boldsymbol{\theta}))=\alpha^*_n$ is the optimal objective value function, and is implicitly related to the problem parameter $\boldsymbol{\theta}$. Since the optimization variable $\alpha$ happens to be the objective function, the desired gradient can be computed by leveraging the Envelope Theorem \cite{milgrom2002envelope}, which enables converting the implicit derivative to an explicit one. Specifically, it states that for a constrained optimization, calculating the total derivative of the optimal objective value function with respect to the problem parameter is equivalent to explicitly evaluating the partial derivative of the Lagrangian with respect to the problem parameter, at the optimal solution:
\begin{align}
\label{func: explicit derivative}
\frac{d V(\boldsymbol{\theta})}{d \boldsymbol{\theta}}=\frac{\partial \mathcal{L}(\mathbf{y}^*, 
\boldsymbol{\lambda}^*,\boldsymbol{\theta})}{\partial \boldsymbol{\theta}}.
\end{align}
This holds because, at the optimal solution, the Lagrangian stationarity condition and the active constraints eliminate implicit derivative terms in $\frac{d V(\boldsymbol{\theta})}{d \boldsymbol{\theta}}$. Thus, we directly implemented $\frac{\partial \mathcal{L}(\mathbf{y}^*, 
\boldsymbol{\lambda}^*,\boldsymbol{\theta})}{\partial \boldsymbol{\theta}}$ in JAX~\cite{jax2018github} to compute the gradient.
\subsection{Differentiating through MPC}\label{sec: diff mpc}
Unlike the previous section, where the implicit gradient of the optimal objective value function with respect to the problem parameter could be obtained through the shortcut, this section involves computing the gradient of the decision variable $\mathbf{\xi}$ with respect to the high-level decision variable $\mathbf{z}$, which is more challenging. We integrate Safe-PDP \cite{safePDP} into our framework to enable differentiation through a state- and control-constrained MPC forward solver. Specifically, in the forward pass, we employ a SQP-based solver~\cite{verschueren2022acados} to optimize the predicted trajectory  
\(\mathbf{\xi}\), ensuring effective satisfaction of all constraints. In the backward pass, we leverage Safe-PDP to compute the gradient  
\(\frac{\partial \mathbf{\xi}}{\partial \mathbf{z}}\).  

To facilitate gradient computation, we first approximate the constrained forward MPC problem with an unconstrained formulation using a logarithmic barrier. Then the discrete-time Pontryagin’s Minimum Principle (PMP) establishes the necessary optimality conditions for the above optimal control problem, and this condition is an implicit function of $\mathbf{x},\mathbf{u},\boldsymbol{\lambda}$ and the high-level decision variable $\mathbf{z}$. By differentiating both sides of the PMP conditions with respect to $\mathbf{z}$, we obtain the \textit{differential PMP} condition that contains the desired gradient $\frac{\partial \mathbf{\xi}}{\partial \mathbf{z}} = \left\{\frac{\partial \mathbf{x}_{0:N}}{\partial \mathbf{z}}, \frac{\partial \mathbf{u}_{0:N-1}}{\partial \mathbf{z}}\right\}$. Notably, this \textit{differential PMP} condition coincides with the backward Riccati recursion of a finite-time LQR problem that has matrix states and control variables $\frac{\partial\mathbf{x}_{k}}{\partial \mathbf{z}}, \frac{\partial \mathbf{u}_{k}}{\partial \mathbf{z}}$.
Thus, solving this finite-horizon LQR via the backward Riccati recursion provides a solution to the \textit{differential PMP} condition and yields $\frac{\partial \mathbf{\xi}}{\partial \mathbf{z}}$.

\begin{algorithm}[]
\caption{Training with Analytical Policy Gradient }\label{alg:close loop train}
\begin{algorithmic}
\STATE {\textbf{Input}}: Drone initial state $\mathbf{x}_{0}$, Gate pose $\mathbf{T}_{\text{gate}\_n}$, Prediction horizon $N$, Episode horizon $H$, Training epoch $K$, Learning rate $\eta$, Batch size $B$.
\STATE {\textbf{Initialization}: $\mathbf\varpi_0$};
\WHILE{$\mathbf{L}$ \text{has not converged} \do}
\FOR{each sample in the batch}
\STATE{Randomly initialize the quadrotor initial position, gate angle, and the goal position};
\FOR{$i = 0,..., H$ \do }
\STATE{\textbf{Forward Pass:}}
\STATE NN inference: $\mathbf{z}_i \leftarrow f_{\mathbf\varpi_K}(\mathbf{o}_i)$;
\STATE MPC optimization: $ \mathbf{\xi}\leftarrow \text{MPC.solve}(\mathbf{x}_i, \mathbf{z}_i) $;
\STATE High-level loss  evaluation: $L_{i} $ using (\ref{prob: high level loss composite});
\STATE State transition: $\mathbf{x}_{i+1}\leftarrow f(\mathbf{x}_i,\mathbf{u}_i)$;   
\STATE{\textbf{Backward Pass:}}
\STATE{ Compute $\frac{\partial L_i}{\partial \mathbf{\xi}_{i}}$ using AD (enabled by Eq.~(\ref{func: explicit derivative}))};
\STATE{ Compute $\frac{\partial \mathbf{\xi}_{i}}{\partial \mathbf{z}_i}$} via the backward Riccati recursion;
\STATE{ Compute $\frac{\partial \mathbf{z}_i}{\partial  \boldsymbol{\varpi}}$ using AD and obtain $\frac{d  L_{i}}{d \mathbf\varpi}$  using (\ref{func: current APG})};
\ENDFOR
\ENDFOR
\STATE Compute the total loss of this episode: $\mathbf{L} \leftarrow \frac{1}{BH}\sum^H_{i=0} L_{i}$;
\STATE Compute the policy gradient:$\frac{\mathrm{d}  \mathbf{L}}{\mathrm{d} \mathbf\varpi}\leftarrow \frac{1}{BH}\sum_{i=0}^{H}\frac{\mathrm{d} L_{i}}{\mathrm{d} \mathbf\varpi}$;
\STATE Update the NN: $\mathbf\varpi_{K+1} \leftarrow \mathbf\varpi_K - \eta \frac{\mathrm{d} \mathbf{L}}{\mathrm{d}\varpi} $;
\ENDWHILE
\end{algorithmic}
\label{alg1}
\end{algorithm}

\section{EXPERIMENTS} \label{sec: experiments}
To demonstrate the advantages of our approach, we perform the following analyses:

\begin{itemize}
\item We compare quadrotor trajectories before and after training, where the initial setting uses a gate-aligned pose and initial fixed weights.
\item We validate our method through extensive hardware experiments and demonstrate improved disturbance rejection capability.
\item We evaluate the training efficiency of our approach against closely related NN-MPC methods and a model-free reinforcement learning baseline.

\end{itemize}

\subsection{Training Setup, Results, and Simulation}\label{sec: simulation}
We keep all configurations in the training and simulation the same as those used for the real deployment, without any additional retuning for sim-to-real transfer.

\subsubsection{Neural networks and training configuration} A multi-layer perceptron (MLP) with two hidden layers of 256 neurons, spectral normalization, and the SiLU activation function is implemented in PyTorch. The observation of the NN, denoted by $\mathbf{o}_i=[\mathbf{x}_i, \mathbf{p}_{\text{goal}},\mathbf{p}_{\text{gate\_corner}\_n}]$, includes the current state of the quadrotor $\mathbf{x}_i$, the goal position $\mathbf{p}_{\text{goal}}$, and the four gate corners positions $\mathbf{p}_{\text{gate\_corner}\_n}$ in the quadrotor body frame $\mathcal{B}$. We choose the Adam optimizer with learning rate $0.0002$ and decay $0.99$ pre $50$ episodes.  $t_{\text{ref}}$ is preset as $1$ s. The high-level loss  weights are $\beta_{\text{gate}}=100, \ \beta_{\text{goal}}=2$, and $ \beta_{\text{control}}=0.001$.

\subsubsection{MPC and high-level decision variable settings} We define the MPC forward and backward problems with CasADi \cite{andersson2019casadi} and acados \cite{verschueren2022acados}, and choose qpOASES as the QP solver. The MPC prediction step is $0.1$ s and the horizon is $20$. The nominal quadrotor dynamics are implemented in the training. The high-level weights variables are given by the output of scaled sigmoid functions at the last layer of the NN. 

The NN is pre-trained to output an initial reference pose at the center of the gate with zero rotation. Benefiting from the use of MPC and Safe-PDP, we are able to impose a hard constraint on the quadrotor’s z-axis position between $0.5$ m and $2$ m, to avoid colliding with the test room ceiling and floor. This constraint is essential in our relatively confined real-world environment.

\subsubsection{Environment configuration}
The gate with size $[0.6, 0.25]$ m is placed at $[0.0, 0.0, 1.9]$ m with a random rotational angle $\theta_{g} \sim \mathcal{U}(-\frac{2\pi}{5},\frac{2\pi}{5})$ rad along the $y$ axis. The quadrotor with collision ellipsoid size $[0.3, 0.3, 0.1]$ m (denoting the diameters along the principal axes) is placed at $\mathbf{p}_{\text{init}}=[0.0, 1.8, 1.2]$ m with goal position $\mathbf{p}_{\text{goal}}=[0.0, -1.8, 1.5]$ m, and both positions are subjected to an uniform random deviation $\mathcal{U}(-0.1,0.1)$ m. 

\subsubsection{Training results in simulation}
\begin{figure}[]
\centering
\vspace{2mm}
    \includegraphics[width=1\columnwidth]
    {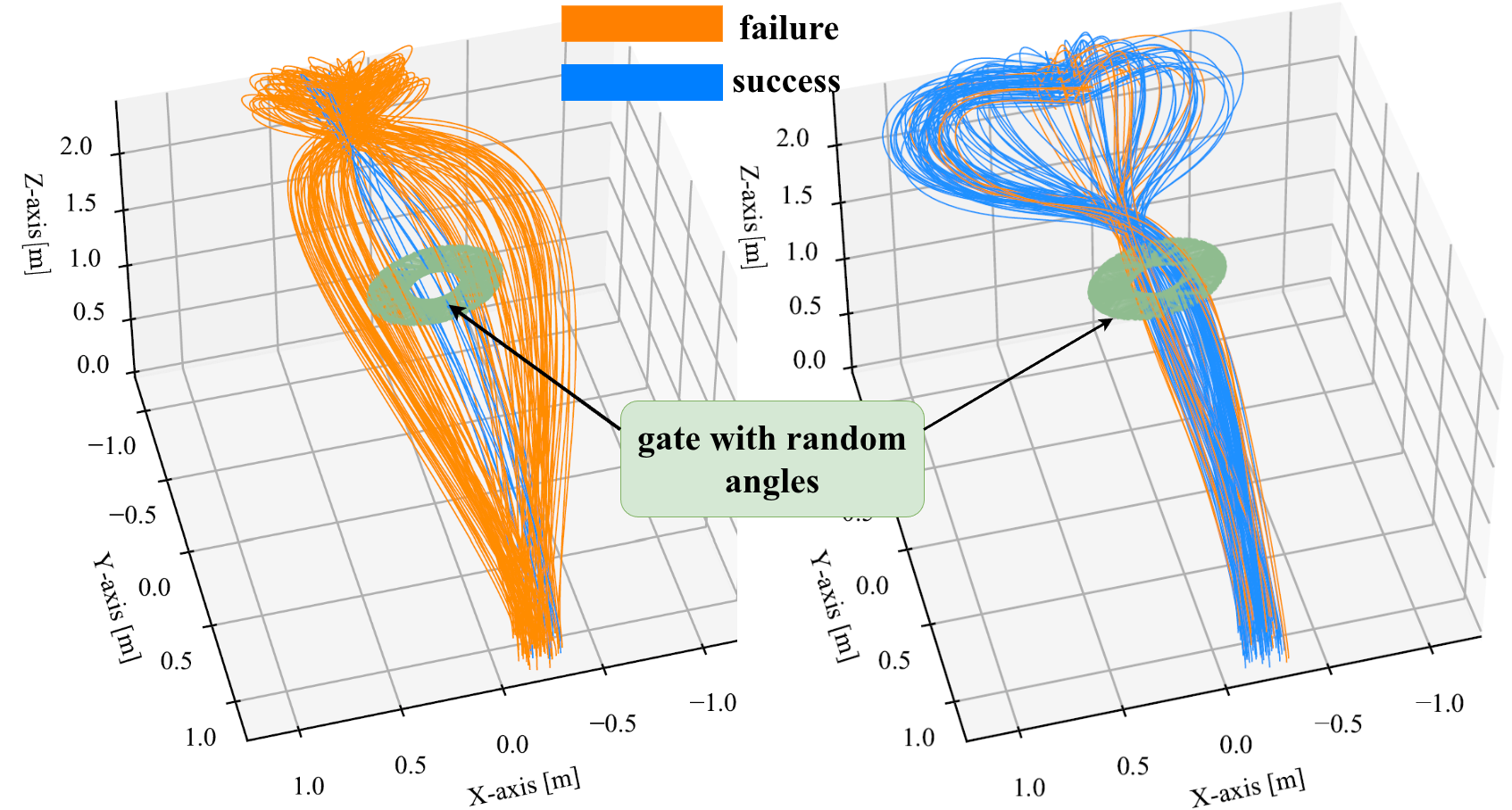}
    \caption{
        Evaluation results of 128 trials before and after training, showing a success rate of 9.38\% and 80.46\%, respectively. 
    } 
    \vspace{-0.4cm}
    \label{fig: training_results_comp}
\end{figure}

We conduct an ablation study to demonstrate the advantage of employing an NN for fast online adaptation of the MPC policy, as shown in Figs.~\ref{fig: training_results_comp}, where blue trajectories indicate successful trials and orange trajectories indicate failures. In Fig.~\ref{fig: training_results_comp} (left), the MPC weights are fixed to the initial outputs of the NN without finetuning and are not optimal, and the reference pose is set to simply match the gate pose, resulting in a very low success rate of 
$9.38 \%$. In contrast, our hybrid framework, trained for only 736k simulation steps, achieves a high success rate of $80.46 \%$, as shown in Fig.~\ref{fig: training_results_comp} (right).

\subsection{Real-world Deployment}\label{sec:real world}
The feasibility of our architecture has been verified in the real world with a custom-designed drone and based on the framework mentioned in \cite{tan2024gestelt}. The drone has a tip-to-tip distance of $25$ cm, and the weight is $0.26$ kg, with a Radxa ZERO 2 pro onboard computer (weight $0.016$ kg, $2.2$ GHz Cortex A73) and both MPC and the NN are deployed onboard, running at $100$ Hz. A body rate controller tracks the desired body rate computed by the MPC at $1000$ Hz. The pose of the quadrotor, provided by the Motion Capture System, is fused with the IMU measurements by an Extended Kalman Filter running at $100$ Hz. Fig. \ref{fig: real flight photos} shows the real flight demonstrations. The quadrotor flies through a narrow gap from multiple approach angles in an agile and precise manner. Moreover, it maintains a minimum clearance of $7.5$ cm while handling gate attitudes from $30^\circ $ to up to $70^\circ $, and achieves peak accelerations of up to $30 \ \mathrm{m/s^2}$. Notably, these agile maneuvers are executed within a highly confined environment offering only $3.6$ m of horizontal and $2$ m of vertical free space.

\begin{figure}[ht]
    \centering
    \includegraphics[width=0.88\columnwidth]{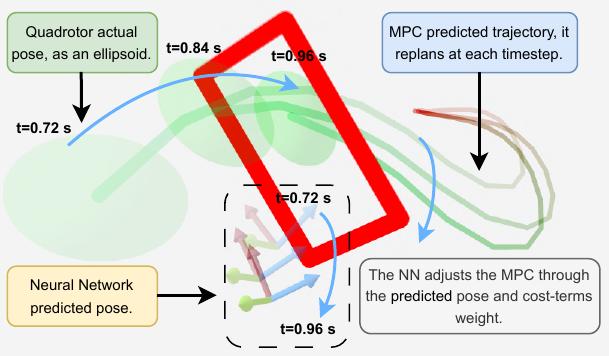}
    \caption{
        Illustration of real flight data during traversal of a gate tilted at $-65^{\circ}$. In real time, the neural network (NN) provides both a high-level reference pose (depicted by coordinate frames) and adaptive cost-term weights for MPC adjustment. The predicted trajectories from MPC and NN-predicted poses at $t=0.72 \ \text{s},0.84 \ \text{s}, \text{and}\ 0.96 \ \text{s}$ are plotted. The complete flight trajectory is captured in Fig.~\ref{fig: real flight photos} D.}
    \label{fig: title_fig}
    \vspace{-0.4cm}
\end{figure}

\subsection{Qualitative Evaluation}
Fig.~\ref{fig: title_fig} shows that the NN-predicted pose $\mathbf{T}_{\text{ref}}$  is adaptively adjusted based on the quadrotor’s state and gate corner position. When an upward deviation of the quadrotor position occurs, the NN compensates this deviation by shifting $\mathbf{T}_{\text{ref}}$ in the opposite direction, ensuring the accurate alignment between the quadrotor and the gate. 

\begin{figure}[]
\centering
    \includegraphics[width=1.00\columnwidth]{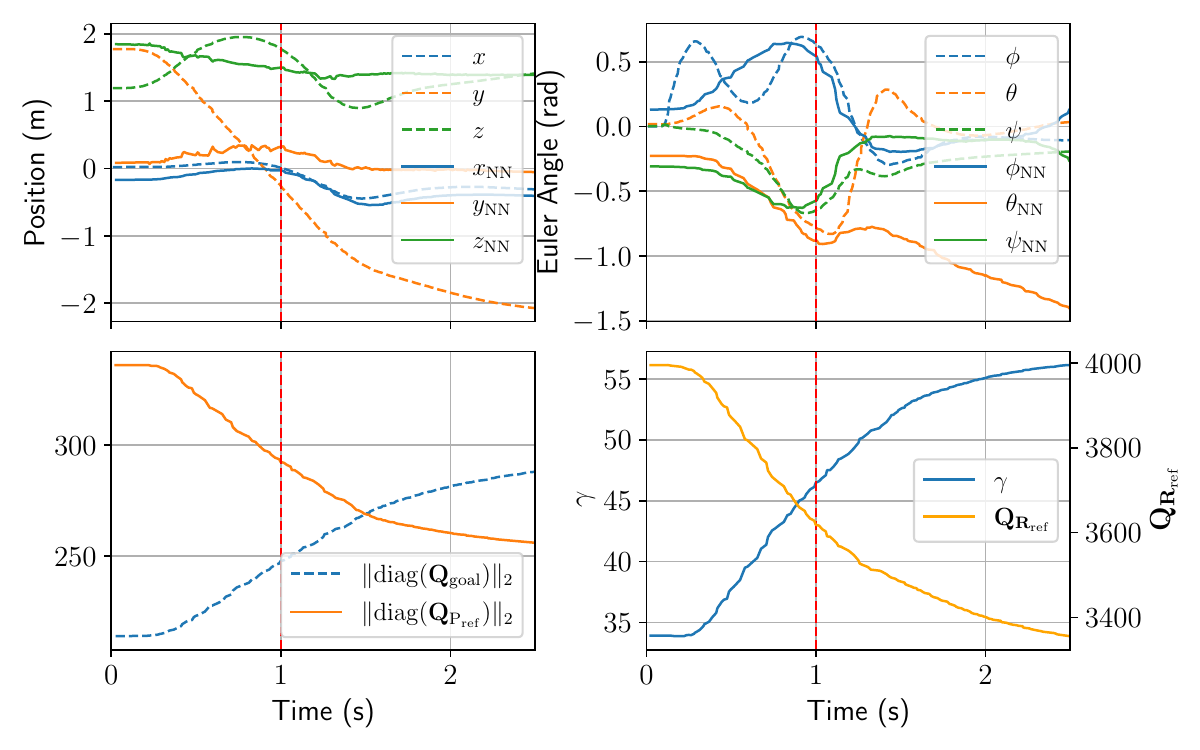}
    \caption{Trajectories of the quadrotor state (dashed) and the NN predicted high-level decision variables (solid) in a real flight. NN predicted position and orientation are dashed lines. $\phi, \theta,\psi$ are the euler angle. The vertical dashed red line indicates the traversal time. } 
    \label{fig: trajectory data vis in axes}
    \vspace{-0.4cm}
\end{figure}

Fig. \ref{fig: trajectory data vis in axes} exhibits the quadrotor state and high-level decision variables trajectories. Initially, the NN prioritizes aligning the quadrotor pose with $\mathbf{T}_{\text{ref}}$ rather than with the goal. This is evident from the higher reference tracking weights compared to the goal-reaching weights. The reference attitude tracking is modulated by two time-varying weights, $\mathbf{Q}_{\mathbf{R}_{\text{ref}}}$ and $\gamma$, governed by the cost function \eqref{eq:c_{ori_vary}}. At the beginning, the combination of a smaller $\gamma$ and a larger $\mathbf{Q}_{\mathbf{R}_{\text{ref}}}$ assigns higher priority to track the attitude component of $\mathbf{T}_{\text{ref}}$.

\begin{figure}[]
\centering
    \includegraphics[width=1\columnwidth]{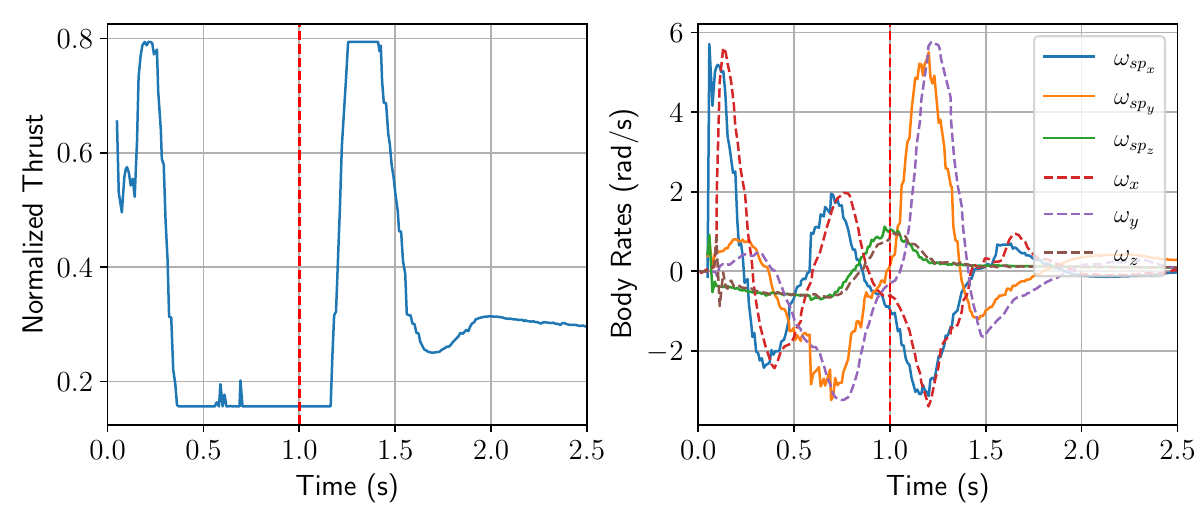}
    \caption{Collective thrust (left) and body rates (right) time-series plots,  in a real flight. On the right, the solid and dashed plots represent body rate setpoints and their measured data, respectively. The vertical dashed red line indicates the traversal time.} 
    \label{fig: control input}
    \vspace{-0.5cm}
\end{figure}

As the quadrotor approaches and passes the gate $\mathbf {T}_{\text{ref}}$, the neural network adapts weight emphasis from pose tracking to goal reaching. Notably, these time-varying, state-dependent weights, along with the reference pose, are difficult to tune manually. Our approach enables automatic tuning through end-to-end learning using analytical optimal policy gradient. Moreover, since high-level decision variables offer meaningful insights into the underlying decision-making process, they provide greater qualitative interpretability than standard RL methods. In addition, to accomplish such an aggressive maneuver in a confined space, the optimized thrust rapidly switches between its upper and lower limits, with peak body rates exceeding $5\ \mathrm{rad/s}$, as shown in Fig.~\ref{fig: control input}. This indicates that the MPC solution operates in a constraint-active regime and fully exploits the available actuation limits.

\subsection{Disturbance Rejection Evaluation}
\begin{figure}[htbp]
\centering
    \includegraphics[width=1\columnwidth]{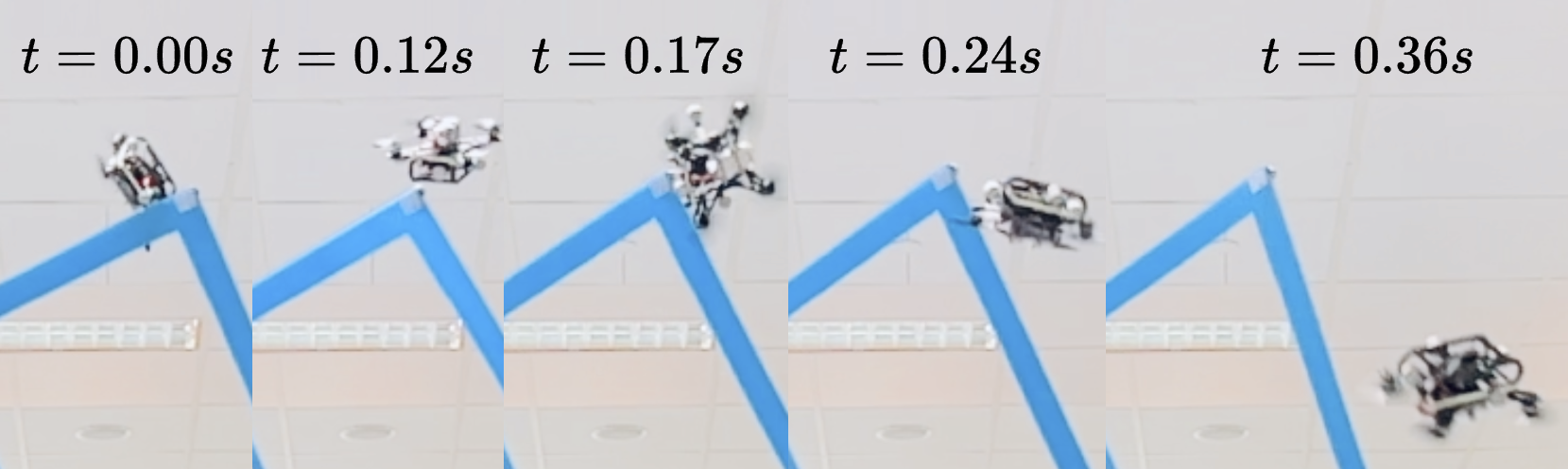}
    \caption{Quadrotor trajectory under collision, caused by inaccurate position estimation. The converged body-rate measurements are illustrated in Fig.~\ref{fig: hit gate control}.}
    \label{fig: hit gate}
    \vspace{-0.4cm}
\end{figure}

\begin{figure}[htbp]
\centering
    \includegraphics[width=1\columnwidth]{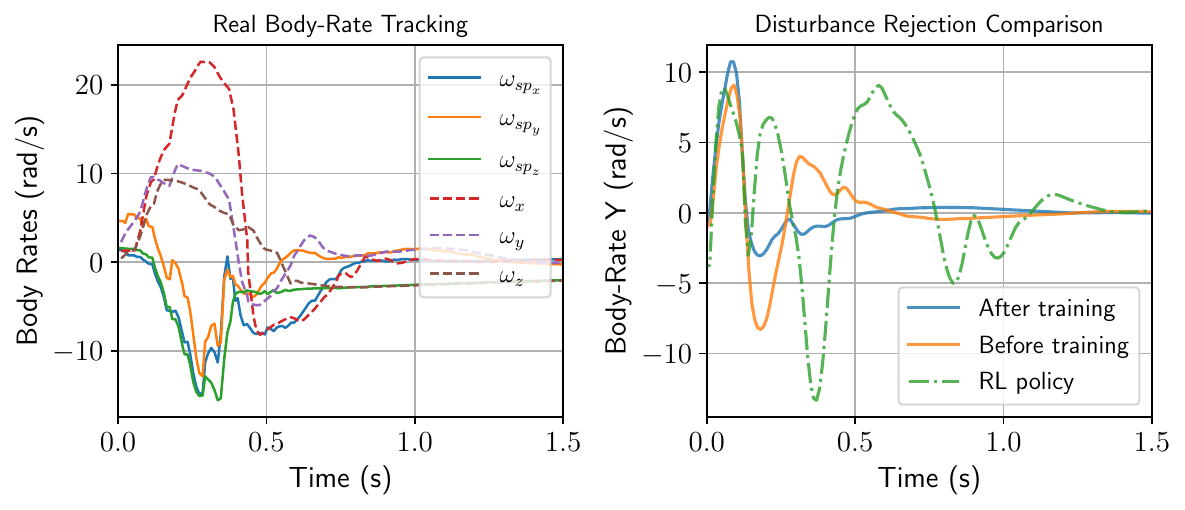}
    \caption{Left: Body-rate time-series plots starting from the collision time during real flight. The solid and dashed curves denote the body-rate setpoints and measured responses, respectively.
    Right: Comparison of the body-rate response along the y-axis under disturbance in simulation for three controllers: MPC before training and after training, and an RL policy.}  
    \label{fig: hit gate control}
    \vspace{-0.4cm}
\end{figure}

Our NN-MPC framework with integrated planning and control demonstrates effective disturbance rejection. A failure case caused by inaccurate position estimation is illustrated in Fig.~\ref{fig: hit gate}, where the quadrotor collides severely with the gate, inducing an extreme body-rate disturbance exceeding $20\ \mathrm{rad/s}$ ($1146 \ \mathrm{deg/s}$). Despite such a severe disturbance, as Fig. \ref{fig: hit gate control} (left) shows, the quadrotor rapidly returns to stable flight in $0.85$ seconds with converged body rate, leveraging online optimization with adaptive cost weights. We first compare the recovery performance in a simulated similar case, from an initial tilt angle of $1\ \mathrm{rad}$ under unseen disturbances, including a $15.5 \ \mathrm{m/s^2}$ linear and an $480\ \mathrm{rad/s^2}$ angular acceleration applied for $0.1\ \mathrm{s}$. The comparison is conducted among our method, a fine-tuned cascaded controller for trajectory tracking \cite{tan2024gestelt} implemented in PX4 SITL, and an end-to-end neural network policy trained with PPO in IsaacLab, all sharing the same simulated body-rate controller. Under these unseen disturbances, both the RL policy and the cascaded controller require longer settling time compared to ours, as summarized in Table~\ref{table: robustness_compare}. We evaluate the effect of training by comparing the body-rate disturbance responses of the proposed MPC before and after training and the RL policy, as shown in Fig.~\ref{fig: hit gate control} (right). The quadrotor controlled by the trained MPC exhibits reduced oscillations and faster recovery, which indicates that our proposed method achieves improved disturbance rejection.
\begin{table}[H]
\centering
\caption{Comparison of the mean settling time averaged over five trials per case.}
\label{table: robustness_compare}
\begin{tabular}{
    >{\centering\arraybackslash}p{0.46\columnwidth} 
    >{\centering\arraybackslash}p{0.46\columnwidth}}
\toprule
Method & Settling Time (s) $\downarrow$ \\
\midrule
NN Policy & 1.30 \\
Cascaded \cite{tan2024gestelt}  & 2.18 \\
Proposed  & \textbf{0.89} \\
\bottomrule
\end{tabular}
\vspace{-0.2cm}
\end{table}


\subsection{Training Efficiency Evaluation}
We first evaluate our proposed method's policy gradient computation time against the method in \cite{wang2023learning, romero2025actor}. We train all policies with 32 parallel environments on an i7-13700 CPU. Since Wang et al. \cite{wang2023learning} implement a two-stage approach, with imitation learning in the second stage, we compare their first stage with ours. For Actor-Critic MPC \cite{romero2025actor}, we measure the policy gradient computation time with different min-batch sizes over 400 mini-batches. Noted that a small mini-batch size leads to noisy gradient estimation. The results are shown in Table \ref{table: iteration_time_compare}. Their methods require much longer times than ours, as our approach derives the analytical gradient for both MPC and the high-level loss and solves fewer optimization problems in each iteration.

We further compare the training efficiency against an end-to-end NN policy trained with a model-free RL method. We employ the PPO algorithm and training in IsaacLab, with the same policy observation and control setup as ours, except for the reward function, which follows \cite{wu2025whole}. We deploy 2048 parallel training environments on an RTX 4090 GPU. The RL policy converges after 200M steps, with $80\%$ success rate. Ours requires only 736k steps to converge, since it leverages the dynamics and numerical optimization with analytical gradient calculation. However, our MPC solver runs on a CPU, which limits parallelization capability, resulting in 70 minutes longer training. The comparison against the RL method is depicted in Table \ref{table: training_eff_compare}. In conclusion, our method achieves a lower computation time of the policy gradient among the NN-MPC approaches and requires fewer training samples compared to the PPO.

\begin{table}[t]
\centering
\caption{Policy gradient compute method and computation time comparison among Wang et al. \cite{wang2023learning}, actor-critic MPC (AC-MPC) \cite{romero2025actor}, and the proposed method, where $M$ refers to mini-batch size.}
\label{table: iteration_time_compare}
\renewcommand{\arraystretch}{1.15}
\setlength{\tabcolsep}{4pt}
\begin{tabularx}{\columnwidth}{
    >{\raggedright\arraybackslash}p{0.29\columnwidth} 
    >{\centering\arraybackslash}p{0.38\columnwidth}
    >{\centering\arraybackslash}p{0.20\columnwidth}
}
\toprule
Method & Grad. Comp. Method & Time (s) $\downarrow$ \\
\midrule
AC-MPC ($M = 8$) &  & 0.22 \\
AC-MPC ($M = 32$)  &  Sampling-based & 0.32 \\
AC-MPC ($M = 128$)   & & 0.58 \\
Wang et al.  & Finite-difference & 0.29 \\
Proposed   & \textbf{Implicit-differentiation } & \textbf{0.16} \\
\bottomrule
\end{tabularx}
\vspace{-0.6cm}
\end{table}

\begin{table}[H]
\centering
\caption{Comparison of the training efficiency between the proposed method and PPO.}
\label{table: training_eff_compare}
\begin{tabularx}{0.42\textwidth}{
  >{\centering\arraybackslash}X
  >{\centering\arraybackslash}X
  >{\centering\arraybackslash}X
  >{\centering\arraybackslash}X
}
\toprule
Method & Sim Steps & Num. of Envs & Training Time \\
\midrule
PPO      & 200M   & 2048 & \textbf{18 min} \\
Proposed & \textbf{736k} & 32 & 93 min \\
\bottomrule
\end{tabularx}
\vspace{-0.2cm}
\end{table}

\section{CONCLUSION}
In this work, we present a hybrid framework that integrates NN and MPC for gate traversal flight. Our framework differentiates through both the MPC and the high-level loss. Leveraging the proposed analytical optimal policy gradient for training, the NN outputs the reference pose and cost-term weights, which act as interpretable control signals for online adaptation of the downstream MPC module. Evidence from both simulation and hardware tests indicates that our method enjoys more efficient training and can precisely plan and control the quadrotor to fly through the narrow gate with effective disturbance rejection. Future work will focus on developing a parallelized optimization solver capable of large-scale training and will subsequently extend the framework to incorporate visual perception for more informed decision-making in unstructured environments.








\bibliographystyle{IEEEtran}
\bibliography{root}

@article{safePDP,
  title={Safe pontryagin differentiable programming},
  author={Jin, Wanxin and Mou, Shaoshuai and Pappas, George J},
  journal={Advances in Neural Information Processing Systems},
  volume={34},
  pages={16034--16050},
  year={2021}
}

@inproceedings{tracy2023differentiable,
  title={Differentiable collision detection for a set of convex primitives},
  author={Tracy, Kevin and Howell, Taylor A and Manchester, Zachary},
  booktitle={2023 IEEE International Conference on Robotics and Automation (ICRA)},
  pages={3663--3670},
  year={2023},
  organization={IEEE}
}

@article{SVD,
  title={An analysis of svd for deep rotation estimation},
  author={Levinson, Jake and Esteves, Carlos and Chen, Kefan and Snavely, Noah and Kanazawa, Angjoo and Rostamizadeh, Afshin and Makadia, Ameesh},
  journal={Advances in Neural Information Processing Systems},
  volume={33},
  pages={22554--22565},
  year={2020}
}

@inproceedings{geist2024learning,
  title={Learning with 3D rotations: a Hitchhiker's guide to SO (3)},
  author={Geist, A Ren{\'e} and Frey, Jonas and Zhobro, Mikel and Levina, Anna and Martius, Georg},
  booktitle={Proceedings of the 41st International Conference on Machine Learning},
  pages={15331--15350},
  year={2024}
}

@inproceedings{tan2024gestelt,
  title={Gestelt: A framework for accelerating the sim-to-real transition for swarm UAVs},
  author={Tan, John and Sun, Tianchen and Lin, Feng and Teo, Rodney and Khoo, Boo Cheong},
  booktitle={2024 IEEE 18th International Conference on Control \& Automation (ICCA)},
  pages={1000--1005},
  year={2024},
  organization={IEEE}
}

@article{verschueren2022acados,
  title={acados—a modular open-source framework for fast embedded optimal control},
  author={Verschueren, Robin and Frison, Gianluca and Kouzoupis, Dimitris and Frey, Jonathan and Duijkeren, Niels van and Zanelli, Andrea and Novoselnik, Branimir and Albin, Thivaharan and Quirynen, Rien and Diehl, Moritz},
  journal={Mathematical Programming Computation},
  volume={14},
  number={1},
  pages={147--183},
  year={2022},
  publisher={Springer}
}

@article{andersson2019casadi,
  title={CasADi: a software framework for nonlinear optimization and optimal control},
  author={Andersson, Joel AE and Gillis, Joris and Horn, Greg and Rawlings, James B and Diehl, Moritz},
  journal={Mathematical Programming Computation},
  volume={11},
  pages={1--36},
  year={2019},
  publisher={Springer}
}

@article{romero2022model,
  title={Model predictive contouring control for time-optimal quadrotor flight},
  author={Romero, Angel and Sun, Sihao and Foehn, Philipp and Scaramuzza, Davide},
  journal={IEEE Transactions on Robotics},
  volume={38},
  number={6},
  pages={3340--3356},
  year={2022},
  publisher={IEEE}
}

@article{song2023reaching,
  title={Reaching the limit in autonomous racing: Optimal control versus reinforcement learning},
  author={Song, Yunlong and Romero, Angel and M{\"u}ller, Matthias and Koltun, Vladlen and Scaramuzza, Davide},
  journal={Science Robotics},
  volume={8},
  number={82},
  pages={eadg1462},
  year={2023},
  publisher={American Association for the Advancement of Science}
}

@article{amos2018differentiable,
  title={Differentiable mpc for end-to-end planning and control},
  author={Amos, Brandon and Jimenez, Ivan and Sacks, Jacob and Boots, Byron and Kolter, J Zico},
  journal={Advances in neural information processing systems},
  volume={31},
  year={2018}
}

@INPROCEEDINGS{bib:Domahidi2013ecos,
author={Domahidi, A. and Chu, E. and Boyd, S.},
booktitle={European Control Conference (ECC)},
title={{ECOS}: {A}n {SOCP} solver for embedded systems},
year={2013},
pages={3071-3076}
}

@article{liu2018search,
  title={Search-based motion planning for aggressive flight in se (3)},
  author={Liu, Sikang and Mohta, Kartik and Atanasov, Nikolay and Kumar, Vijay},
  journal={IEEE Robotics and Automation Letters},
  volume={3},
  number={3},
  pages={2439--2446},
  year={2018},
  publisher={IEEE}
}

@inproceedings{lee2010geometric,
  title={Geometric tracking control of a quadrotor UAV on SE (3)},
  author={Lee, Taeyoung and Leok, Melvin and McClamroch, N Harris},
  booktitle={49th IEEE conference on decision and control (CDC)},
  pages={5420--5425},
  year={2010},
  organization={IEEE}
}

@article{song2022policy,
  title={Policy search for model predictive control with application to agile drone flight},
  author={Song, Yunlong and Scaramuzza, Davide},
  journal={IEEE Transactions on Robotics},
  volume={38},
  number={4},
  pages={2114--2130},
  year={2022},
  publisher={IEEE}
}

@article{romero2025actor,
  title={Actor--Critic Model Predictive Control: Differentiable Optimization Meets Reinforcement Learning for Agile Flight},
  author={Romero, Angel and Aljalbout, Elie and Song, Yunlong and Scaramuzza, Davide},
  journal={IEEE Transactions on Robotics},
  volume={42},
  pages={673--692},
  year={2025},
  publisher={IEEE}
}

@article{hartley2013rotation,
  title={Rotation averaging},
  author={Hartley, Richard and Trumpf, Jochen and Dai, Yuchao and Li, Hongdong},
  journal={International journal of computer vision},
  volume={103},
  pages={267--305},
  year={2013},
  publisher={Springer}
}

@article{WANG2022GCOPTER,
    title={Geometrically Constrained Trajectory Optimization for Multicopters}, 
    author={Wang, Zhepei and Zhou, Xin and Xu, Chao and Gao, Fei}, 
    journal={IEEE Transactions on Robotics}, 
    year={2022}, 
    volume={38}, 
    number={5}, 
    pages={3259-3278}, 
    doi={10.1109/TRO.2022.3160022}
}

@inproceedings{wang2023learning,
  title={Learning agile flight maneuvers: Deep {SE(3)} motion planning and control for quadrotors},
  author={Wang, Yixiao and Wang, Bingheng and Zhang, Shenning and Sia, Han Wei and Zhao, Lin},
  booktitle={2023 IEEE International Conference on Robotics and Automation (ICRA)},
  pages={1680--1686},
  year={2023},
  organization={IEEE}
}

@article{xie2023learning,
  title={Learning agile flights through narrow gaps with varying angles using onboard sensing},
  author={Xie, Yuhan and Lu, Minghao and Peng, Rui and Lu, Peng},
  journal={IEEE Robotics and Automation Letters},
  volume={8},
  number={9},
  pages={5424--5431},
  year={2023},
  publisher={IEEE}
}

@inproceedings{wu2025whole,
  title={Whole-body control through narrow gaps from pixels to action},
  author={Wu, Tianyue and Chen, Yeke and Chen, Tianyang and Zhao, Guangyu and Gao, Fei},
  booktitle={2025 IEEE International Conference on Robotics and Automation (ICRA)},
  pages={11317--11324},
  year={2025},
  organization={IEEE}
}

@inproceedings{falanga2017aggressive,
  title={Aggressive quadrotor flight through narrow gaps with onboard sensing and computing using active vision},
  author={Falanga, Davide and Mueggler, Elias and Faessler, Matthias and Scaramuzza, Davide},
  booktitle={2017 IEEE international conference on robotics and automation (ICRA)},
  pages={5774--5781},
  year={2017},
  organization={IEEE}
}

@software{jax2018github,
  author = {James Bradbury and Roy Frostig and Peter Hawkins and Matthew James Johnson and Chris Leary and Dougal Maclaurin and George Necula and Adam Paszke and Jake Vander{P}las and Skye Wanderman-{M}ilne and Qiao Zhang},
  title = {{JAX}: composable transformations of {P}ython+{N}um{P}y programs},
  url = {http://github.com/jax-ml/jax},
  version = {0.3.13},
  year = {2018},
}

@inproceedings{neunert2016fast,
  title={Fast nonlinear model predictive control for unified trajectory optimization and tracking},
  author={Neunert, Michael and De Crousaz, C{\'e}dric and Furrer, Fadri and Kamel, Mina and Farshidian, Farbod and Siegwart, Roland and Buchli, Jonas},
  booktitle={2016 IEEE international conference on robotics and automation (ICRA)},
  pages={1398--1404},
  year={2016},
  organization={IEEE}
}

@article{milgrom2002envelope,
  title={Envelope theorems for arbitrary choice sets},
  author={Milgrom, Paul and Segal, Ilya},
  journal={Econometrica},
  volume={70},
  number={2},
  pages={583--601},
  year={2002},
  publisher={Wiley Online Library}
}

@inproceedings{ferede2025one,
  title={One net to rule them all: Domain randomization in quadcopter racing across different platforms},
  author={Ferede, Robin and Blaha, Till and Lucassen, Erin and De Wagter, Christophe and De Croon, Guido CHE},
  booktitle={2025 IEEE International Conference on Robotics and Automation (ICRA)},
  pages={6357--6363},
  year={2025},
  organization={IEEE}
}

@article{hu2025narrow,
  title={Narrow Gap Traversing via Differentiable Physics},
  author={Hu, Yu and Zou, Danping},
  journal={Journal of Shanghai Jiaotong University (Science)},
  pages={1--8},
  year={2025},
  publisher={Springer}
}

\end{document}